\documentclass[11pt]{article}

\usepackage[final]{acl}
\usepackage{booktabs}

\usepackage{times}
\usepackage{latexsym}
\usepackage{multirow}
\usepackage{tcolorbox}
\tcbuselibrary{breakable}  
\usepackage{amsmath}  
\usepackage{amsfonts} 
\usepackage{makecell}

\usepackage[T1]{fontenc}

\usepackage[utf8]{inputenc}

\usepackage{microtype}
\usepackage{amsmath}
\usepackage{tcolorbox}
\tcbuselibrary{skins,breakable}
\usepackage{inconsolata}

\usepackage{graphicx}
\usepackage{enumitem}
\title{Seeing is Believing: Aligning Prompt Rewriting with Visual Anchors for Text-to-Image Generation}

\author{
 \textbf{Xuanyi Liu \textsuperscript{1}}
 \textbf{Deyi Ji\textsuperscript{2}}
 \textbf{Junyu Lu\textsuperscript{3}}
 \textbf{Jing Wang\textsuperscript{1}}
  \textbf{Lanyun Zhu\textsuperscript{4}}
 \textbf{Qianxiong Xu\textsuperscript{5}}\\
 \textbf{Xuhang Chen\textsuperscript{6}}
 \textbf{Tianrun Chen\textsuperscript{7}}
 \textbf{Siwei Ma \textsuperscript{1}}
\\
 \textsuperscript{1}Peking University
 \textsuperscript{2}Tencent
 \textsuperscript{3}Dalian University of Technology
 \textsuperscript{4}Tongji University \\
 \textsuperscript{5}Nanyang Technological University
 \textsuperscript{6}University of Cambridge
 \textsuperscript{7}Zhejiang University
\\
}

\begin{document}
\maketitle

\begingroup
\renewcommand{\thefootnote}{}
\footnotetext{Xuanyi Liu completed this work while interning at Tencent as part of the Tencent Rhino-Bird Research Elite Program, with Deyi Ji as the program leader. Correspondance to: Deyi Ji<jideyi16@foxmail.com>, Siwei Ma<swma@pku.edu.cn>}
\addtocounter{footnote}{-1}
\endgroup

\begin{abstract}
Despite the impressive capabilities of text-to-image (T2I) models, an intent–generation gap often persists due to the brevity and ambiguity of user prompts. Existing approaches primarily polish the prompt for fluency and readability. However, the enhancement process still lacks visual grounding. 
As a result, the rewriter may over-infer missing details, causing an intent-generation gap.
To address this limitation, we propose FaithRewriter, a novel prompt-enhancement framework for T2I generation. 
Specifically, FaithRewriter first leverages a multimodal MLLM to generate an image from the original prompt as an intermediate visual cue. 
This cue is then combined with the prompt and fed into a large-scale LLM to produce visually grounded augmentations that better reflect how the intended content should appear in images. 
Finally, these augmentations are distilled into a small-scale LLM for efficient deployment, enhancing its ability to generate effective T2I prompts.
Experiments show that FaithRewriter yields prompts that are more faithful to the user intent and more visually plausible than strong baselines, helping narrow the intent-generation gap.
\end{abstract}
\section{Introduction}

Text-to-image (T2I) models can generate high-quality images from natural language~\cite{peebles2023scalable,rombach2022high,Cao_2026}. However, users rarely provide the precise, detailed prompts these models expect. Instead, ambiguous or underspecified inputs force generators to guess spatial layouts, physical interactions, and implicit constraints. This creates an \textit{intent--generation gap}: the resulting images may look realistic, but they often fail to faithfully reflect the user's true intent.

\begin{figure}[t]
  \centering
  \includegraphics[width=0.95\linewidth]{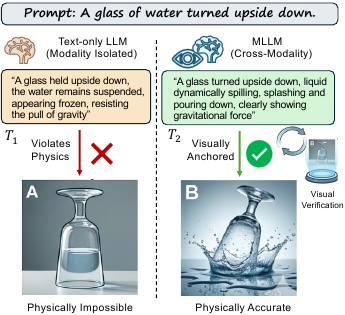}
  \caption{Comparison between standard rewriting and FaithRewriter. Standard rewriting ($T_1$) often introduces hallucinations due to modal isolation. FaithRewriter utilizes a Visual Anchor to diagnose these errors and produces a verified revision ($T_2$).}
  \label{fig:fig_intro}
  \vspace{-0.05in}
\end{figure}

A common solution is prompt rewriting, where a Large Language Model (LLM) expands the initial input into a richer instruction. While this improves linguistic fluency, it ignores \textit{visual faithfulness}. Because text-only rewriters~\cite{promptenhancer,wu2025reprompt} operate entirely in language space, they often hallucinate details that sound linguistically reasonable but are visually or physically impossible~\cite{wang2026everything}. For example (Figure~\ref{fig:fig_intro}), rewriting ``a glass of water turned upside down'' might yield a fluent description of water suspended in mid-air, which violates basic physics. Here, blind textual expansion does not enhance the prompt; it pollutes the original intent.


We argue that faithful prompt rewriting should not be a simple text-expansion task, but rather an alignment with visual reality. The core problem is that text-only models lack visual common sense. Even if a prompt reads perfectly, it can easily produce an unrealistic image with floating objects, wrong counting, or broken physics. Therefore, the challenge is: how can we teach a text-only model to ``see'' and avoid visual mistakes during rewriting?

Our core idea is simple: the best way to catch a text hallucination is to actually draw it. When we render a text-only rewrite into an image, hidden logical errors instantly become obvious visual failures. We call this intermediate image a \textit{Visual Anchor}. By generating a Visual Anchor, we can use a Multimodal LLM (MLLM) as an automated visual debugger. The MLLM looks at the image, pinpoints exactly where the layout or physics went wrong, and writes a corrected prompt to fix the specific error. It then verifies that a second generated image actually fixes the problem without losing the user's original request. This ensures the final prompt is not just a longer sentence, but a visually verified instruction.

We introduce \textbf{FaithRewriter}, a ``Simulate-Then-Distill'' framework. Because generating images and querying MLLMs is too slow for real-time inference, we only use Visual Anchors to build our training data offline. For thousands of prompts, we run the diagnosis and correction process described above. This gives us highly valuable preference pairs: a ``bad'' rewrite (fluent but visually flawed) and a ``good'' rewrite (visually corrected and verified). Crucially, the ``bad'' rewrite serves as a perfect \textit{hard negative} to prevent the model from learning superficial tricks like simply writing longer sentences. We distill this knowledge into a compact text-only model using Supervised Fine-Tuning (SFT) and Direct Preference Optimization~\cite{rafailov2024directpreferenceoptimizationlanguage} (DPO). Then the text rewriter learns to avoid visual pitfalls without generating any images during inference.

To evaluate this, we introduce \textbf{FaithT2I}, a benchmark featuring information-dense prompts and VQA-style questions targeting spatial, physical, and semantic constraints. Extensive experiments across various benchmarks demonstrate that FaithRewriter is effective.


Our contributions are threefold:
\begin{itemize}[leftmargin=*]
    \item We identify the core failure of standard prompt rewriting: a description can be linguistically perfect but visually impossible.
    
    \item We propose FaithRewriter, a Simulate-Then-Distill framework that uses Visual Anchors to expose flawed rewrites and construct verified revisions for preference learning.
    
    \item We introduce FaithT2I-test and show that SFT+DPO distillation transfers cross-modal priors into an efficient text-only rewriter with consistent gains across benchmarks and generators.
\end{itemize}
\section{Related Work}
\begin{figure*}
    \centering
    \includegraphics[width=1.0\linewidth]{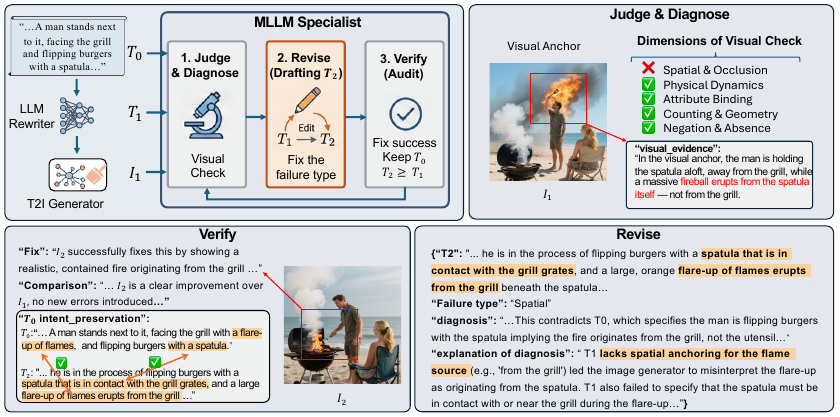}
    \caption{Overview of FaithRewriter. Given an original prompt $T_0$, a text-only rewriter produces $T_1$, which is rendered into $I_1$. An MLLM diagnoses the visual failure, revises $T_1$ into a grounded prompt $T_2$, and verifies that the resulting image $I_2$ fixes the error while preserving the original intent. DPO training is omitted for clarity.}
    \label{fig:method}
\end{figure*}

\subsection{Prompt Rewriting}
Prompt rewriting transforms short or underspecified user instructions into more detailed prompts~\cite{promptenhancer, beautifulprompt,prewrite,2024instances,zhou2025learning,li2024learning,manas2024improving}. Existing approaches often treat discrete prompts as parameters to be optimized. For instance, PRewrite \cite{prewrite} trains a rewriter LLM using downstream task rewards, allowing the model to explore and identify more effective prompts. Similarly, \cite{zhou2025learning} employs a meta-LLM iteratively to revise the prompt for a given test instance based on the task LLM's output. \cite{li2024learning} uses a powerful LLM to generate rewrites and distills this capability into a compact model. Furthermore, works like PromptEnhancer\cite{promptenhancer} focus on generating task-specific or personalized prompts, often for NLP tasks like text classification or personalized email generation. However, they primarily operate within the \textit{textual} modality, lacking an explicit mechanism to ensure the \textit{visual plausibility} of the generated text, which is crucial for text-to-image (T2I) synthesis.

\subsection{Text-to-Image Synthesis}

The field of text-to-image (T2I) synthesis \cite{esser2024scaling,han2024evalmuse,hao2023optimizing,podell2023sdxl,tao2023galip} has advanced significantly, with models like HunyuanImage and Qwen-Image achieving high fidelity through advanced architectures. However, research has primarily focused on enhancing the model's internal capabilities, often overlooking a critical external factor: the quality of the input prompt itself. The prompt serves as the sole bridge between human intent and the generative process. Ambiguous prompts force even powerful models to guess details. This limitation is evident in benchmarks like T2I-CompBench \cite{huang2023t2i}, which show models struggle with complex object relations and negations—precisely where human descriptions are most prone to ambiguity. Addressing this requires shifting focus from making models understand vague language better to making language clearer for models. Recent work explores this interface: Hunyuan Enhancer trains an LLM to rewrite prompts, improving output alignment, while approaches using rich text or structured formatting provide more explicit signals to guide generation.
\section{Method}
\subsection{Problem Formulation}
\label{sec:problem_formulation}
Let $x$ be the original user prompt and $y$ be its rewrite. A text-only rewriter models a distribution $p_\text{text}(y | x)$, favoring linguistically fluent prompts. However, T2I generation requires a different criterion: the rewritten prompt should produce an image that is faithful to the intended visual content of x. We denote the ideal visually grounded distribution as $p_\text{cross}(y | x)$.

The key mismatch is that textual likelihood does not guarantee visual realizability, yielding:
\begin{equation}
    P_{\text{text}}(y|x) \to \text{high}, \quad \text{while} \quad P_{\text{cross}}(y|x) \approx 0.
\end{equation}
Such a rewrite is a textual hallucination: fluent in text but harmful for generation. Although it may appear specific and coherent, it can cause visual errors such as misplaced objects, wrong attribute binding, implausible dynamics, missing negated objects, or incorrect counts and relations.

FaithRewriter exposes these hidden failures through rendering. Given a candidate rewrite $y$, a frozen T2I model generates an image $I$, which reveals the visual consequences of $y$. A multimodal judge then checks whether $I$ satisfies the original prompt $x$. This enables us to build training pairs where the rejected rewrite is not merely textually weak, but directly linked to a visual failure.

\subsection{FaithRewriter}
\label{sec:pipe}
FaithRewriter consists of two stages. 

The first stage constructs visually grounded preference pairs through simulation, diagnosis, revision, and verification. The second stage distills these pairs into a compact text-only rewriter using SFT and DPO. This design uses expensive multimodal reasoning only during offline data construction, while keeping inference as efficient as standard text-only rewriting.

\subsubsection{Visual Failure Externalization}

As Figure~\ref{fig:method}, given an original prompt $T_0$, we first generate a baseline rewrite $T_1$ using a text-only LLM. $T_1$ is intentionally allowed to be fluent and detailed, because our goal is not to compare against a weak negative, but to expose the specific failure modes of text-only expansion. We then render $T_1$ using a frozen T2I generator to obtain an intermediate image $I_1$, which we call the visual anchor. The visual anchor transforms a latent textual risk into an observable image-level outcome.

The visual anchor serves two purposes. First, it externalizes latent prompt hallucinations. Errors that were difficult to detect in text, such as an ambiguous flame source or an incorrect occlusion relation, become directly observable in the image. Second, it provides concrete evidence for revision. Instead of asking an MLLM to improve the prompt abstractly, we ask it to explain what went wrong in $I_1$ and which part of $T_1$ caused the failure. This evidence-grounded revision process reduces the risk of arbitrary rewriting and makes each correction traceable to a specific visual failure.

\subsubsection{Structured Diagnosis and Revision}

Given the original prompt $T_0$, the initial rewrite $T_1$, and the rendered image $I_1$, the MLLM diagnoses whether $T_1$ leads to a visual failure in $I_1$. We consider six common failure types: spatial logic and occlusion, attribute binding, physical dynamics, counting and geometry, negation and absence, and implicit concepts. These categories cover the main cases where a prompt can remain fluent in language but become ambiguous or implausible after visual instantiation.

For each sample, the MLLM first checks whether $I_1$ clearly violates the intent of $T_0$. Samples without a clear failure are discarded to avoid subjective or unnecessary edits. For failed samples, the MLLM identifies the failure type, points to the visual evidence in $I_1$, and explains which part of $T_1$ causes the error. This step ensures that the revision is driven by an explicit causal link between the rewritten text and the observed image failure.

The MLLM then generates a revised prompt $T_2$. The revision follows two constraints. First, it should fix the diagnosed failure by adding explicit visual grounding, such as object positions, attachment relations, source-target bindings, or physical causality. Second, it should preserve the original intent of $T_0$. In particular, the MLLM is instructed not to remove secondary objects, change attributes, alter the scene context, or introduce unrelated details when fixing the primary error. These constraints encourage targeted correction rather than unconstrained prompt expansion.

\subsubsection{Cycle Verification}
A revised prompt may fix the diagnosed error while introducing new deviations from the original intent. Therefore, simply generating $T_2$ is insufficient; we need to verify whether the correction is visually realized after rendering. To this end, we perform cycle verification by rendering $T_2$ with the same T2I generator to obtain $I_2$. A strict MLLM judge then compares $T_0$, $I_1$, and $I_2$ using three binary criteria: fix verification, whether $I_2$ resolves the failure diagnosed in $I_1$; intent preservation, whether $I_2$ preserves the original constraints and entities in $T_0$; and anti-regression, whether $I_2$ avoids new visual errors while fixing the original one. This verification step turns revision from an open-ended rewriting process into a closed-loop quality control process grounded in the rendered outcome.

Only when all three checks pass do we accept the pair, recorded as $T_\text{win} = T_2$ and $T_\text{lose} = T_1$. Otherwise, the sample is discarded. This strict filtering improves preference-data reliability by ensuring that the positive prompt is not only linguistically improved, but also visually validated. It also ensures that $T_\text{lose}$ is a true hard negative: fluent and detailed, yet linked to an observed visual failure. As a result, the retained pair isolates visual grounding as the primary difference between the preferred and rejected rewrites, making it suitable for preference optimization.


\subsubsection{Counterfactual Preference Distillation}

After constructing the verified preference dataset, we train a compact text-only rewriter. We first apply supervised fine-tuning on positive pairs $(T_0, T_\text{win})$, which teaches the student the syntax and style of visually grounded prompts. However, SFT alone does not explicitly teach the model to reject plausible but visually harmful details. This motivates a preference-learning stage that contrasts correct visual grounding against matched but flawed alternatives.

We therefore apply DPO using matched hard-negative pairs $(T_\text{win} > T_\text{lose})$. This design is important. A naive preference pair such as $T_\text{win} > T_0$ would confound visual faithfulness with prompt length, because $T_\text{win}$ is typically longer and more detailed than the raw prompt. In contrast, $T_\text{win}$ and $T_\text{lose}$ are both expanded rewrites of the same $T_0$. They are similar in fluency, verbosity, and style, but differ in whether their added details are visually grounded. DPO therefore learns a more targeted preference: prefer visually realizable details over visually implausible ones.

\begin{figure*}[t]
    \centering
    \includegraphics[width=1.0\linewidth]{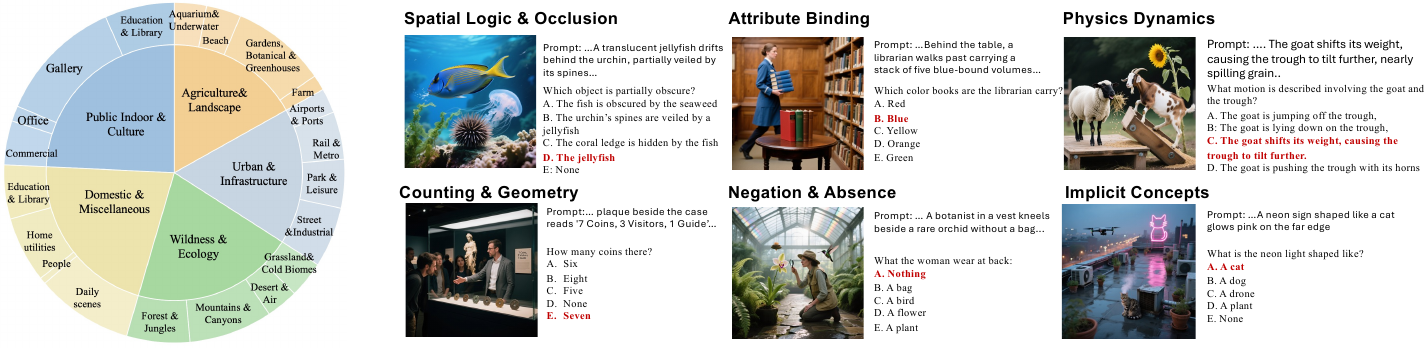}
    \caption{Overview of FaithT2I-test: scene distribution and example prompt--question--answer pairs.}
    \label{fig:dataset}
\end{figure*}

The DPO objective is:
\begin{equation}
\begin{aligned}
    \mathcal{L}_{\text{DPO}}(& \theta)  = -  \mathbb{E}_{(T_0, T_{\text{win}},  T_{\text{lose}}) \sim \mathcal{D}} \Bigg[ \log \sigma \left( \right. \\
    & \beta \log \frac{\pi_{\theta}(T_{\text{win}}|T_0)}{\pi_{\text{ref}}(T_{\text{win}}|T_0)} 
    - \beta \log \frac{\pi_{\theta}(T_{\text{lose}}|T_0)}{\pi_{\text{ref}}(T_{\text{lose}}|T_0)} \left. \right) \Bigg],
\end{aligned}
\end{equation}
where $\pi_{\text{ref}}$ is the SFT-initialized model, and $\beta$ controls the deviation margin. This optimization forces the student to look beyond surface-level language patterns, explicitly unlearning the tendency to generate textual hallucinations and pushing its distribution toward the cross-modal ideal $p_{\text{cross}}$. 

After distillation, the student performs faithful prompt rewriting using only text input,
while the simulation and verification process remains offline.


\subsection{FaithT2I Benchmark}

Unlike generic metrics, FaithT2I systematically evaluates whether a generated image faithfully satisfies dense visual constraints. As shown in Figure~\ref{fig:dataset}, we classify potential T2I failures into six macro-categories: (1) Spatial logic \& occlusion, which focuses on relative layouts and obstruction; (2) Attribute binding, which evaluates color, size, and property associations; (3) Physical dynamics, which tests causality and contact; (4) Counting \& geometry, which examines counts, ratios, and grids; (5) Negation \& absence, which checks correct omission; and (6) Implicit concepts, which evaluates commonsense visual interpretation. 


\noindent\textbf{Prompt and Question Construction.}
To ensure diversity, we prompt an MLLM to synthesize dense prompts ($T_0$) across diverse real-world domains, as shown in Figure~\ref{fig:dataset}. Each sample pairs $T_0$ with 10 VQA-style questions to evaluate the generated image. To prevent forced guessing in cases of severe misalignment, each question includes an explicit ``none of the above'' option. This design encourages the evaluator to measure concrete visual satisfaction rather than rewarding generic text-image similarity. Computational cost is in Appendix~\ref{app:data_cost}.

\noindent\textbf{Verification and Reliability.}
For dataset reliability, a subset of 300 samples is manually verified by three human annotators, yielding substantial agreement with a Cohen's Kappa of 0.72. To avoid evaluation leakage, we completely decouple the evaluator models, including GPT-4o, Gemini, and Qwen3-VL, from the training-time textual prompt generation model DeepSeek-R1~\cite{Guo_2025}. This separation helps ensure that the benchmark evaluates general visual faithfulness rather than memorized preferences of a single model family.

\section{Experiment}
\subsection{Experiment Setup}

\textbf{Benchmarks.} To evaluate FaithRewriter, we utilize both our curated dataset and public benchmarks.
FaithT2I (Ours): Constructed via the pipeline in Section~\ref{sec:pipe}, this dataset focuses on hard cases for T2I generation. We curate splits for training (\textit{FaithT2I-3K}, \textit{FaithT2I-10K}, \textit{FaithT2I-100K}) and test on the held-out test set \textit{FaithT2I-Test}, and other public benchmarks like SpatialGenEval~\cite{wang2026everything}, a benchmark containing complex spatial prompts. We also evaluate on DPG-Bench~\cite{hu2024ella} and GenEval~\cite{ghosh2023geneval} to verify generalization capabilities.



\noindent\textbf{Baselines \& Models.} 
For the rewriter backbone, we experiment with LLaMA3.1-8B, Hunyuan-7B-Instruct, and Qwen3-8B-Base who have similar size. We conduct on these comparisons: (1) $T_0$ (Raw): Direct user prompts; (2) Zero-shot: directly generate (3) Text-modal: Fine-tuned on text-only rewritten data ($T_{1}$) (4) FaithRewriter: Fine-tuned on cross-modality rewritten data ($T_{2}$)
For image generation, we employ Qwen-Image~\cite{qwenimage} as default and compare with Hunyuan-Image 2.1~\cite{hunyuanimage}, and FLUX.1-dev.
We further include two existing prompt rewriting baselines: PromptEnhancer~\cite{promptenhancer}, built upon Hunyuan-7B-Instruct, and RePrompt~\cite{wu2025reprompt} . For fairness, we adapt RePrompt from its original Qwen2.5-3B setting to the same Qwen3-8B backbone used in our experiments.

\noindent\textbf{Implementation Details.} See in Appendix~\ref{appendix:implementation}.

\begin{table*}[t]
\centering
\scalebox{0.73}{
\begin{tabular}{lccccccccccc}
\toprule
\multirow{2}{*}{Method}                                         & 
\multicolumn{3}{c}{VQA Scoring~$\uparrow$}  & \multicolumn{3}{c}{Textual Alignment Scoring~$\uparrow$}                                       & \multicolumn{3}{c}{Preference Ranking~$\downarrow$}                                      & \multicolumn{1}{c}{\multirow{2}{*}{CLIP~$\uparrow$}} & \multicolumn{1}{c}{\multirow{2}{*}{FID~$\downarrow$}}  \\
 & \multicolumn{1}{c}{Gemini} & \multicolumn{1}{c}{GPT4o} & \multicolumn{1}{c}{Qwen3VL} & \multicolumn{1}{c}{Gemini} & \multicolumn{1}{c}{GPT4o} & \multicolumn{1}{c}{Qwen3VL} & \multicolumn{1}{c}{GPT4o} & \multicolumn{1}{c}{Gemini} & \multicolumn{1}{c}{Qwen3VL} & \multicolumn{1}{c}{} & \\ \midrule
$T_0$ (raw)  & 62.12 & 60.28 & 64.39 & 3.303 & 3.939 & 3.994 & - & - & - & 0.752 & 23.94 \\ 

\midrule
\multicolumn{12}{l}{\textit{\textbf{LLaMa3.1-8B}}} \\
Zeroshot  & 55.00 & 58.23 & 60.32 & 3.314 & 3.421 & 3.456 & 2.20 & 2.22 & 2.25 & 0.762 & 23.76 \\ 
Text modal  & 59.12 & 62.35 & 64.52 & 3.229 & 3.689 & 3.977 & 2.09 & 1.95 & 2.07 & 0.774 & 23.58 \\
FaithRewriter & \textbf{63.27} & \textbf{65.61} & \textbf{70.65} & \textbf{4.037} & \textbf{4.249} & \textbf{4.403} & \textbf{1.71} & \textbf{1.83} & \textbf{1.68} & \textbf{0.784} & \textbf{23.41}  \\ 
\midrule
\multicolumn{12}{l}{\textit{\textbf{Hunyuan-7B-Instruct}}}  \\
PromptEnhancer  & 68.59 & 72.69 & 74.43 & 3.472 & 3.491 & 3.483 & 2.24 & \textbf{1.89} & 2.28 & \textbf{0.813} & \textbf{22.84} \\ 
Zeroshot  & 64.18 & 65.82 & 62.94 & 3.327 & 3.364 & 3.381 & 3.11 & 3.39 & 3.07 & 0.768 & 23.67 \\ 
Text modal & 66.37 & 68.15 & 67.06 & 3.176 & 3.863 & 3.925 & 2.81 & 2.77 & 2.71 & 0.796 & 23.25 \\ 
FaithRewriter & \textbf{70.83} & \textbf{73.41} & \textbf{77.94} & \textbf{3.949} & \textbf{4.001} & \textbf{4.164} & \textbf{1.84} & 1.95 & \textbf{1.94} & 0.808 & 22.87 \\ \midrule
\multicolumn{12}{l}{\textit{\textbf{Qwen3-8B-Base}}}  \\
RePrompt  & 69.15 & 68.91 & 76.12 & 3.194 & 3.857 & 3.891 & 2.23 & 1.97 & 1.89 & 0.821 & 22.58 \\ 
Zeroshot  & 59.73 & 61.34 & 70.83 & \textbf{4.122} & 3.745 & 3.773 & 2.29 & 2.38 & 2.63 & 0.762 & 23.76 \\ 
Text modal & 68.78 & 69.40 & 75.65 & 3.263 & 3.814  & 3.963 & 2.19 & 2.03 & 1.82 & 0.828 & 22.63  \\
FaithRewriter  & \textbf{74.80} & \textbf{73.71} & \textbf{78.91} & 4.075 & \textbf{4.098} & \textbf{4.392} & \textbf{1.52} & \textbf{1.59} & \textbf{1.55} & \textbf{0.847} & \textbf{22.19}  \\ 
\bottomrule
\end{tabular}}
\caption{Main results on FaithT2I-test.}
\label{tab:main_faith}
\end{table*}

\begin{table*}[t]
\centering
\scalebox{0.7}{\begin{tabular}{lccccccccccc}
\toprule
\multirow{2}{*}{Method} & 
\multirow{2}{*}{Overall} & \multicolumn{2}{c}{Spatial Foundation} & \multicolumn{3}{c}{Spatial Perception} & \multicolumn{3}{c}{Spatial Reasoning} & \multicolumn{2}{c}{Spatial Interaction} \\ \cmidrule(lr){3-4} \cmidrule(lr){5-7} \cmidrule(lr){8-10} \cmidrule(lr){11-12}
 &  & Object & Attribute & Position & Orientation & Layout & Comparison & Proximity & Occlusion & Motion & Causal \\ \midrule
$T_0$ (raw) & 63.09 & 87.16 & 80.47 & 69.23 & 66.24 & 73.64 & 24.42 & 66.81 & 38.22 & 65.25 & 59.42 \\ 
Zeroshot & 64.58 & 87.55 & 80.12 & 71.10 & 66.80 & 74.80 & 27.20 & 68.20 & 39.85 & 68.10 & 62.10 \\ 
Text modal & 66.23 & 87.72 & 80.47 & 72.79 & 67.95 & 76.34 & 29.97 & 71.79 & 40.92 & 70.23 & 64.11 \\
FaithRewriter  & \textbf{70.83} & \textbf{88.44} & \textbf{82.60} & \textbf{74.78} & \textbf{73.36} & \textbf{80.33} & \textbf{35.66} & \textbf{76.20} & \textbf{53.44} & \textbf{73.93} & \textbf{69.52} \\ 
\bottomrule
\end{tabular}}
\caption{Main results on SpatialGenEval. By utilizing visual anchors, our Cross-modal enhancer demonstrates a comprehensive leap over text-only baselines, particularly in difficult sub-domains like \textit{Occlusion} and \textit{Comparison} .}
\label{tab:main_spatial}
\end{table*}

\begin{table*}[!hbtp]
\centering
\scalebox{0.65}{\begin{tabular}{l|cccccc|ccccccc}
\toprule
\multirow{2}{*}{Enhancer} & 
\multicolumn{6}{c|}{DPG-Bench} & \multicolumn{7}{c}{GenEval} \\
& Global & Entity & Attribute & Relation & Other & Overall & Single Object & Two Object & Counting & Colors & Position & Attribute & Overall \\ 
\midrule
$T_0$ (raw)    & 91.32 & 91.56 & 92.02 & 94.31 & 92.73 & 88.32 & \textbf{0.99} & 0.92 & 0.89 & 0.88 & 0.76 & 0.77 & 0.87  \\
Zeroshot       & 91.47 & \textbf{91.63} & 92.14 & 94.82 & 92.59 & 88.74 & 0.98 & 0.91 & 0.89 & 0.88 & 0.77 & 0.77 & 0.87 \\
Text modal     & 91.85 & 91.70 & 92.25 & 95.12 & 92.41 & 89.21 & 0.98 & 0.91 & 0.90 & \textbf{0.89} & 0.77 & 0.78 & 0.88 \\
FaithRewriter  & \textbf{92.64} & 91.48 & \textbf{93.15} & \textbf{96.53} & \textbf{93.08} & \textbf{90.34} & \textbf{0.99} & \textbf{0.94} & \textbf{0.93} & 0.87 & \textbf{0.80} & \textbf{0.79} & \textbf{0.89} \\ 
\bottomrule
\end{tabular}}
\caption{Main results on public benchmarks (DPG-Bench and GenEval). FaithRewriter (Cross modal) achieves a robust +2\% overall improvement on both benchmarks. \textbf{Bold} indicates the best performance.}
\label{tab:main_dpg}
\end{table*}

\subsection{Main results}
\subsubsection{Results on FaithT2I-Test}

We first evaluate different prompt enhancement methods on the FaithT2I-10K dataset, with results summarized in Table~\ref{tab:main_faith}.


Overall, FaithRewriter consistently achieves the best performance across nearly all metrics and backbones. Compared with text-only rewriting, the cross-modal variant yields higher VQA scores, stronger textual alignment, better human-preference-style ranking from multiple MLLM judges, and more favorable CLIP/FID results. These improvements are especially consistent across Gemini 3 Pro~\cite{comanici2025gemini}, GPT-4o~\cite{hurst2024gpt}, and Qwen3-VL~\cite{bai2025qwen3} evaluators, showing that the gains are not tied to a single judge. At the same time, all rewriting-based methods outperform the raw prompt baseline, confirming the value of prompt enhancement itself, while the comparison between Text modal and Cross-modal further shows that visual-anchor-based supervision provides additional benefits beyond pure textual fine-tuning.

\begin{table*}[t]
\centering

\label{tab:module_ablation}
\resizebox{\textwidth}{!}{%
\begin{tabular}{lccc|cccccccc}
\toprule
\multirow{2}{*}{\textbf{Method Variants}} & \multicolumn{3}{c|}{\textbf{Components}} & \multicolumn{3}{c}{\textbf{VQA Scoring} $\uparrow$} & \multicolumn{3}{c}{\textbf{Textual Alignment Scoring} $\uparrow$} & \multirow{2}{*}{\textbf{CLIP} $\uparrow$} & \multirow{2}{*}{\textbf{FID} $\downarrow$} \\
& Visual Anchor & Cycle Verif. & DPO & Gemini & GPT-4o & Qwen3VL & Gemini & GPT-4o & Qwen3VL & & \\
\midrule
Text-only Rewriting              & $\times$ & $\times$ & $\times$    & 66.37 & 68.15 & 67.06 & 3.176 & 3.863 & 3.925 & 0.796 & 23.25 \\
Text-only Self-Refinement        & $\times$ & $\times$ & $\times$    & 66.52 & 68.29 & 67.28 & 3.284 & 3.882 & 3.948 & 0.797 & 23.19 \\
Image-Cond. Revision (No Filter) & $\checkmark$ & $\times$ & $\times$ & 68.14 & 70.63 & 71.55 & 3.512 & 3.914 & 4.015 & 0.799 & 23.06 \\
Cycle Verif. (SFT Only)          & $\checkmark$ & $\checkmark$ & $\times$ & 69.61 & 72.18 & 75.31 & 3.765 & 3.957 & 4.089 & 0.803 & 22.95 \\
\midrule
\textbf{FaithRewriter (Ours)} & $\checkmark$ & $\checkmark$ & $\checkmark$ & \textbf{70.83} & \textbf{73.41} & \textbf{77.94} & \textbf{3.949} & \textbf{4.001} & \textbf{4.164} & \textbf{0.808} & \textbf{22.87} \\
\bottomrule
\end{tabular}}
\caption{Ablation study on the necessity of simulation and framework components based on Hunyuan-7B-Instruct.}
\label{tab:module_ablation}
\end{table*}

\subsubsection{Results on SpatialGenEval}

We further evaluate our method on SpatialGenEval, and the results are shown in Table~\ref{tab:main_spatial}. Compared with the raw, zeroshot, and text-only baselines, the cross-modal model achieves the strongest overall performance and improves almost all sub-dimensions of spatial intelligence. The gains are particularly clear in more challenging categories such as \textit{Comparison}, \textit{Occlusion}, \textit{Motion}, and \textit{Causal}, where text-only enhancement remains limited. These results suggest that FaithRewriter improves general prompt faithfulness, and is effective for prompts requiring precise spatial reasoning, physical interaction, and visually grounded logic.

\subsubsection{Results on DPG-Bench and GenEval}

We further evaluate FaithRewriter on two public benchmarks, DPG-Bench and GenEval. As shown in Table~\ref{tab:main_dpg}, our cross-modal variant consistently outperforms both the raw prompt baseline and the text-only rewriting baseline on both benchmarks, yielding around 2\% overall improvement. Although this experiment is not our main focus, it provides additional evidence that the gains of FaithRewriter can transfer to general-purpose public benchmarks.




\subsection{Ablation Study}

\subsubsection{Component Ablation}
To isolate the contribution of each component in FaithRewriter, we compare the full method with several degraded variants in Table~\ref{tab:module_ablation}. 
Text-only Rewriting removes visual feedback entirely, while Text-only Self-Refinement allows the LLM to iteratively critique and refine its own prompt only in the textual domain. 
Image-Conditioned Revision introduces the visual anchor but removes strict cycle verification, retaining all revised prompts. 
Cycle Verification (SFT Only) uses only the verified grounded prompts for SFT, while Full FaithRewriter further applies DPO with matched hard negatives.

The comparison shows three main findings. 
First, Text-only Self-Refinement brings only marginal improvement over standard Text-only Rewriting, suggesting that textual critique alone is insufficient for detecting spatial and physical hallucinations. 
Second, introducing visual anchors substantially improves performance, confirming that rendering intermediate images is important for externalizing latent prompt failures. 
Third, cycle verification and DPO provide complementary gains: cycle verification improves the reliability of the training data, while DPO further teaches the student to prefer visually grounded rewrites over linguistically similar but visually flawed alternatives. Additional SFT/DPO ablation on Qwen-8B is in Appendix~\ref{appendix:more_results}.

\subsubsection{Robustness Across T2I Models}
We evaluate the universality of FaithRewriter by applying the same enhanced prompts to three diverse T2I models: FLUX, Hunyuan Image, and Qwen-Image. As shown in Table~\ref{tab:abl_t2i}, FaithRewriter consistently boosts visual assessment scores across all generators compared to the raw prompts ($T_0$) and intra-modal baselines. 
This demonstrates that FaithRewriter learns model-agnostic visual features effectively bridging the intent gap.

\begin{table}[!htbp]
\centering
\scalebox{0.7}{\begin{tabular}{l|cccc}
\toprule
\multirow{2}{*}{Methods} & \multirow{2}{*}{T2I Model} & \multicolumn{3}{c}{VQA Scoring~$\uparrow$} \\
& & Gemini & GPT-4o & Qwen3VL \\ 
\midrule
$T_0$ (raw) & FLUX-dev1 & 55.12 & 54.38 & 58.21 \\
FaithRewriter & FLUX-dev1 & \textbf{67.25} & \textbf{65.92} & \textbf{70.43} \\ \midrule
$T_0$ (raw) & Hunyuan Image & 60.15 & 58.94 & 62.15 \\
FaithRewriter & Hunyuan Image & \textbf{72.84} & \textbf{71.56} & \textbf{76.54} \\ \midrule
$T_0$ (raw) & Qwen-Image & 62.12 & 60.28 & 64.39 \\ 
FaithRewriter & Qwen-Image & \textbf{74.80} & \textbf{73.71} & \textbf{78.91} \\
\bottomrule
\end{tabular}}
\caption{Ablation study on different T2I backbones.}
\label{tab:abl_t2i}
\end{table}





\subsubsection{Data Scaling and Generalization}
We investigate the scaling laws of our method by training on a 3k, 10k, 100k dataset. As shown in Table~\ref{tab:abl_data_scaling}, the model trained on 100k samples consistently outperforms the 3k and 10k version. The performance gap between the 100k model and the text-only baseline suggests that our cross-modal distillation process benefits significantly from increased data diversity. The results indicate that the advantage of FaithRewriter is not saturated, showing strong potential for further improvement with larger-scale cross-modal simulation.

\begin{table}[t]
\centering
\scalebox{0.8}{\begin{tabular}{l|ccc}
\toprule
\multirow{2}{*}{Configuration} & \multicolumn{3}{c}{Textual Alignment Scoring~$\uparrow$} \\
 & Gemini & GPT-4o & Qwen3VL \\ 
\midrule
$T_0$ (raw) & 3.303 & 3.939 & 3.994 \\
\midrule
FaithRewriter (3k) & 3.712 & 3.924 & 4.105 \\
FaithRewriter (10k) & 4.075 & 4.098 & \textbf{4.392} \\
FaithRewriter (100k) & \textbf{4.150} & \textbf{4.123} & 4.241 \\ 
\bottomrule
\end{tabular}}
\caption{Ablation study on training data scaling. Performance gains correlate with the size of training data.}
\label{tab:abl_data_scaling}
\end{table}





\begin{figure*}[t]
    \centering
    \includegraphics[width=1.0\linewidth]{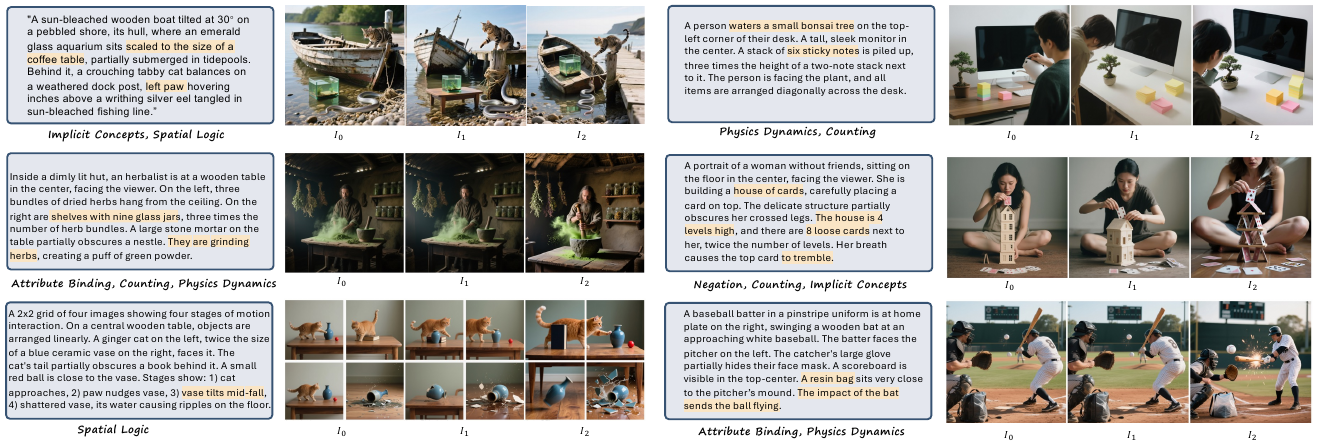}
    \caption{Exemplar qualitative comparisons across diverse evaluation dimensions (e.g., spatial logic, physical dynamics, negation, and quantitative counting). $I_0$, $I_1$, and $I_2$ denote images generated by the raw prompt ($T_0$), the standard text-only rewrite ($T_1$), and our FaithRewriter prompt ($T_2$), respectively.}
    \label{fig:case}
\end{figure*}

\subsection{User Study}
\begin{table}[!htbp]
\centering
\scalebox{0.72}{\begin{tabular}{l|ccccc}
\toprule
\multirow{2}{*}{Configuration} & \multicolumn{3}{c}{User Preference Score~$\downarrow$} & \multirow{2}{*}{Overall~$\downarrow$} & \multirow{2}{*}{Rank} \\
 & LLaMA & Hunyuan & Qwen & & \\ 
\midrule
$T_0$(raw) & 3.65 & 3.72 & 3.29 & 3.55 & 4 \\
Zeroshot & 2.63 & 2.45 & 2.95 & 2.68 & 3 \\
Text modal & 2.29 & 2.23 & 2.57 & 2.36 & 2 \\
FaithRewriter & \textbf{1.43} & \textbf{1.60} & \textbf{1.19} & \textbf{1.41} & 1 \\ 
\bottomrule
\end{tabular}}
\caption{User study results.}
\label{tab:user_study}
\end{table}
We conduct a large-scale user study with 62 participants and 60 samples from generated data. Given the original prompt $T_0$ and outputs from three configurations ($T_0$, Zeroshot, Text modal, and Cross-modal), participants rank the images according to how faithfully they satisfy the prompt. The results in Table~\ref{tab:user_study} show a clear preference for Cross-modal, which achieves the best overall mean ranking.

\noindent\textbf{VLM-Human Correlation.} Taking Qwen-VL as an example, we measure the correlation between VLM preference scores and averaged human rankings across the study samples, obtaining a Pearson correlation of 0.76. This suggests that VLM judgments are reasonably aligned with human.

\subsection{Qualitative Case Analysis}
\label{sec:case_analysis}

We present qualitative comparisons in Figure~\ref{fig:case}. 


\noindent\textbf{Resolving Semantic Ambiguity and Implicit Concepts.}
Text-only rewriters often drift when handling metaphorical or implicit concepts. In the ``House of cards'' case (Figure~\ref{fig:case}, Middle-Right), the prompt refers to a structure made of playing cards, but both $I_0$ and $I_1$ misinterpret it as a miniature wooden house. In contrast, FaithRewriter ($I_2$) correctly renders a four-level card structure.  
A similar issue appears in the ``Boat'' case (Top-Left), where the aquarium is described as \textit{``scaled to the size of a coffee table''}. The text-only baseline $I_1$ hallucinates a coffee table, while $I_2$ preserves the intended scale without introducing unwanted objects.

\noindent\textbf{Enforcing Quantitative and Geometric Constraints.}
T2I models often fail on precise counting and relative geometry. In the ``Sticky notes'' case (Top-Right), the prompt requires one stack of six notes to be \textit{``three times the height''} of a two-note stack. Both $I_0$ and $I_1$ miss this relation, whereas $I_2$ satisfies the count and height constraints.  
In the ``Motion Grid'' case (Bottom-Left), the prompt requires a strict \textit{2$\times$2} layout showing four sequential stages. Only $I_2$ follows this structure faithfully, while $I_1$ produces a disorganized composition.

\noindent\textbf{Restoring Physical Dynamics and Causality.}
Text-only rewriters may enrich action descriptions linguistically but still fail to induce correct visual dynamics. In the ``Baseball'' case (Bottom-Right), $I_0$ and $I_1$ show a mostly static scene and miss the \textit{``impact of the bat''}. By contrast, $I_2$ depicts a more dynamic impact with visible physical consequences.  
This also appears in the ``Herb grinding'' case (Middle-Left), where only $I_2$ correctly visualizes the \textit{``puff of green powder''}.

\section{Conclusion}

In this work, we identified the inherent limitation of text-only prompt rewriting methods, which often produce visually implausible descriptions due to a lack of cross-modal anchor. To address this, we introduced FaithRewriter, a novel framework built on a "Simulate-Then-Distill" paradigm. FaithRewriter leverages a powerful MLLM as a cognitive simulator to generate visually-anchored prompt enhancements and then efficiently distills this capability into a compact, text-only LLM. We build a corresponding benchmark FaithT2I-test. Extensive experimental results demonstrate that FaithRewriter significantly outperforms existing methods in generating faithful and realistic prompts, effectively bridging intent-generation gap. 


\section*{Limitations}

\noindent\textbf{Offline Computational Overhead.} A primary limitation of our framework lies in the computational cost of the data construction phase. The ``Simulate'' step requires rendering intermediate images via a diffusion model and invoking a large-scale Multimodal LLM (MLLM) for structured diagnosis and cycle-verification. Consequently, scaling the training dataset (e.g., to millions of pairs) is resource-intensive and time-consuming. However, it is crucial to note that this overhead is strictly confined to the offline training phase. Thanks to our distillation strategy, the deployed student model operates entirely in the text modality, ensuring high inference-time efficiency without requiring visual encoders or online rendering.

\noindent\textbf{Boundaries of the Native Generative Prior.} The effectiveness of any prompt rewriting method, including FaithRewriter, is inherently bounded by the intrinsic capabilities of the downstream T2I generator. If the base generative model entirely lacks the prior knowledge of a specific concept (e.g., an extremely rare object) or suffers from hard architectural bottlenecks (e.g., accurately rendering illegible text or generating exactly 15 identical objects), an enhanced prompt cannot magically inject these missing capabilities. While FaithRewriter expertly reorganizes instructions to maximize the generator's potential, it cannot overcome fundamental rendering deficits.

\bibliography{custom}

\newpage
\section{Appendix}
\label{sec:appendix}


\subsection{Implementation Details.} 
\label{appendix:implementation}




We evaluate FaithRewriter with three student rewriter backbones: LLaMa-3.1-8B,Qwen3-8B, and Hunyuan-7B-Dense-Instruct. For Qwen and LLaMA-family backbones, we implement full-parameter fine-tuning with LLaMA-Factory. Training is conducted on one node with 8 NVIDIA H20 GPUs under CUDA 12.3, cuDNN 9, PyTorch 2.3, and DeepSpeed ZeRO-3 optimization. The maximum sequence length is set to 2048, and training is performed in bfloat16 precision with FlashAttention-2 enabled.

For supervised fine-tuning, we train for 3 epochs with a per-device batch size of 1, gradient accumulation of 2, a learning rate of $1\times10^{-5}$, a cosine learning-rate scheduler, and a warmup ratio of 0.1. For Hunyuan-7B-Dense-Instruct, we use the Angel-PTM training framework with full-parameter fine-tuning. The sequence length is set to 32768, the global batch size is 16, the micro batch size is 1, and the model is trained for 5 epochs with a learning rate of $5\times10^{-5}$ and a minimum learning rate of $1\times10^{-5}$.

For DPO training, we initialize from the corresponding SFT checkpoint and use full-parameter optimization with the standard sigmoid DPO loss. Unless otherwise specified, we train for 3 epochs with a per-device batch size of 1, gradient accumulation of 8, a learning rate of $5\times10^{-6}$, a cosine scheduler, a warmup ratio of 0.1, and $\beta=0.1$. All variants based on the same backbone use identical infrastructure and hyperparameters, differing only in the supervision data and optimization objective.

\subsection{Data Construction Cost and Statistics}
\label{app:data_cost}

We provide additional details on the offline cost of constructing visually grounded preference data. 
The construction pipeline consists of four main steps: initial text-only rewriting, visual-anchor rendering, MLLM-based diagnosis and revision, and cycle verification. 
For each candidate prompt, the pipeline requires one rendering for the initial rewrite $T_1$ and one additional rendering for the revised prompt $T_2$ if the sample enters the verification stage. 
The MLLM is called for structured diagnosis, prompt revision, and final verification.

In our implementation, we first generated a larger pool of candidate prompts and then applied confidence gating and cycle verification to retain high-quality preference pairs. 
Approximately 55--65\% of candidate samples were identified as containing clear visual failures after the initial visual-anchor diagnosis. 
After rendering the revised prompt and applying the final cycle-verification checks, around 35--45\% of the original candidates were retained as verified preference pairs. 
This filtering process removes cases where the initial image has no clear failure, where the revision changes the original intent, or where the revised rendering introduces new visual errors.

\begin{table*}[h]
\centering
\small
\begin{tabular}{lccc}
\toprule
Stage & Main Operation & Approx. Calls per Candidate & Retained Ratio \\
\midrule
Initial rewriting & Generate $T_1$ from $T_0$ & 1 LLM call & 100\% \\
Visual anchoring & Render $I_1$ from $T_1$ & 1 T2I call & 100\% \\
Diagnosis and revision & Diagnose $I_1$ and generate $T_2$ & 1 MLLM call & 55--65\% \\
Cycle verification & Render $I_2$ and verify improvement & 1 T2I call + 1 MLLM call & 35--45\% \\
\bottomrule
\end{tabular}
\caption{Approximate cost profile of the FaithRewriter data construction pipeline. 
The ratios are computed after each filtering stage.}
\label{tab:data_cost}
\end{table*}

For the 10K-scale training split, this corresponds to roughly 22K--28K candidate prompts before filtering, depending on the random seed and generator behavior. 
The full construction process requires approximately two T2I renderings and two MLLM calls for each candidate that reaches the final verification stage. 
We do not include monetary cost because API prices and deployment settings vary across providers and time. 
Instead, we report the number of model calls, which provides a provider-independent estimate of construction cost.

Importantly, this cost is incurred only once during offline data construction. 
After distillation, the final FaithRewriter model is a compact text-only rewriter and does not require T2I rendering or MLLM-based verification at inference time.

\subsection{Additional Impact of Training Stages}
\label{appendix:more_results}
Table~\ref{tab:abl_training} isolates the contributions of each training stage in our ``Simulate-Then-Distill'' paradigm on Qwen-8B-Base model.

\noindent\textbf{SFT Efficiency:} The SFT-only model achieves a substantial performance leap over the baseline ($T_0$), confirming that distilling the MLLM's ``visual syntax'' provides a strong initialization.

\noindent\textbf{DPO Refinement:} The addition of the DPO stage yields further consistent gains across all judges. This validates the effectiveness of our hard negative mining strategy. By explicitly penalizing the linguistically fluent but visually flawed $y_{\text{lose}}$, DPO sharpens the model's judgment, reducing hallucinations that SFT alone cannot eliminate.

\begin{table}[!htbp]
\centering
\scalebox{0.8}{\begin{tabular}{l|ccc}
\toprule
\multirow{2}{*}{Configuration} & \multicolumn{3}{c}{Textual Alignment Scoring~$\uparrow$} \\
 & Gemini & GPT4o & Qwen3VL \\ \midrule
T0 (raw) & 3.303 & 3.939 & 3.994 \\
Cross-modal (SFT) & 3.843 & 3.825 & 3.879\\
Cross-modal (SFT+DPO) & \textbf{4.075} & \textbf{4.098} & \textbf{4.392} \\ 
\bottomrule
\end{tabular}}
\caption{Ablation on training based on Qwen-8B.}
\label{tab:abl_training}
\end{table}


\subsection{VQA Data example}
\label{sec:appendix_dataset}

To systematically evaluate the spatial intelligence and cross-modal grounding capabilities of T2I models, our FaithT2I benchmark contains diverse, information-dense prompts coupled with omni-dimensional VQA pairs. 
Below, we present representative examples from the dataset. Each card illustrates the prompt, the scene context, and the 10 fine-grained questions targeting specific failure modes (e.g., spatial logic, attribute binding, physical dynamics, and shape-based object disambiguation). The inclusion of the ``E: None'' option prevents evaluators from forced-guessing when the generated image completely fails to follow the prompt.

\begin{tcolorbox}[
    enhanced, breakable,
    colback=blue!2!white, colframe=black!75,
    coltitle=white, colbacktitle=black!75,
    fonttitle=\bfseries,
    title={Example 1: FaithT2I VQA Dataset },
    boxrule=0.8pt, arc=2pt,
    left=6pt, right=6pt, top=4pt, bottom=4pt
]
\small
\textbf{Prompt:} In a dim art gallery hallway, a photographer on the right moves leftward with a camera raised. On the left, a curator adjusts a painting, facing the photographer. The photographer has one lens; the curator holds three placards. The photographer’s reflection glimmers slightly in a nearby glass case, blurred by motion.
\tcblower

\textbf{Omni-dimensional VQA Pairs:}
\begin{itemize}
    \setlength{\itemsep}{2pt}
    \item What object categories are explicitly mentioned besides people? \\
    \textit{Options:} (A) A lens and placards, (B) A tripod and a frame, (C) A flashlight and a notebook, (D) A sculpture and a bench, (E) None \\
    \textbf{Answer: A}
    
    \item  What are the described attributes of the gallery hallway? \\
    \textit{Options:} (A) Bright and noisy, (B) Dim and quiet, (C) Crowded and colorful, (D) Empty and dark, (E) None \\
    \textbf{Answer: B}

    \item From a third-person perspective, where is the photographer located? \\
    \textit{Options:} (A) On the left side, (B) In the center, (C) On the right side, (D) Behind the glass case, (E) None \\
    \textbf{Answer: C}

    \item  What is the facing orientation of the curator? \\
    \textit{Options:} (A) Facing away, (B) Facing the painting, (C) Facing right, (D) Facing the photographer, (E) None \\
    \textbf{Answer: D}
    
    \item How are the photographer and the curator arranged in the scene? \\
    \textit{Options:} (A) One behind the other, (B) On opposite sides, (C) Next to each other, (D) In a diagonal line, (E) None \\
    \textbf{Answer: B}

    \item  How does the number of placards compare to the number of lenses? \\
    \textit{Options:} (A) Fewer placards than lenses, (B) Same number, (C) More placards than lenses, (E) None \\
    \textbf{Answer: C}
    
    \item Given the photographer is moving leftward toward the curator, which person is the photographer getting closer to? \\
    \textit{Options:} (A) A visitor, (B) The curator, (C) A security guard, (D) A gallery assistant, (E) None \\
    \textbf{Answer: B}
    
    \item  According to the scene, is any object described as being occluded or partially hidden? \\
    \textit{Options:} (A) Face hidden by camera, (B) Placards hidden by body, (C) No occlusion mentioned, (D) Glass case hides reflection, (E) None \\
    \textbf{Answer: C}
    
    \item  What is the described motion interaction between the photographer and the curator? \\
    \textit{Options:} (A) Moving away from each other, (B) Photographer moving toward curator, (C) Curator walking toward photographer, (D) Standing still, (E) None \\
    \textbf{Answer: B}
    
    \item  What causes the photographer’s reflection to appear blurred in the glass case? \\
    \textit{Options:} (A) Dim lighting, (B) Photographer’s motion, (C) Dirty glass, (D) Placards blocking view, (E) None \\
    \textbf{Answer: B}
\end{itemize}
\end{tcolorbox}

\subsection{Diagnosis Examples}

We provide details of the diagnosis output of the image we used in the main pipeline.

\begin{tcolorbox}[
    enhanced, breakable,
    colback=gray!5!white, colframe=black!75,
    coltitle=white, colbacktitle=black!75,
    fonttitle=\bfseries,
    title={Diagnosis Example},
    boxrule=0.8pt, arc=2pt,
    left=6pt, right=6pt, top=4pt, bottom=4pt
]

\textbf{\textcolor{blue}{[Phase 1: Modal Gap Input]}}
\begin{itemize}
    \item \textbf{User Intent ($T_0$):} \textit{...The man leans back slightly from a flare-up of flames.}
    \item \textbf{Text-only Rewrite ($T_1$):} \textit{...he is in the process of flipping burgers with a spatula but leans back exaggeratedly from a flare-up of large, orange flames...}
\end{itemize}

\textbf{\textcolor{red}{[Phase 2: Structured Diagnosis \& Revision]}}
\begin{itemize}
    \item \textbf{Visual Evidence (from $I_1$):} In $I_1$, the man is holding the spatula aloft, away from the grill, while a massive fireball erupts from the spatula itself---not from the grill. This contradicts $T_0$.
    \item \textbf{Diagnosis:} \textbf{Physics Failure.} The fire originates from the spatula, not the grill, and the man's posture is incorrect. $I_1$ depicts a surrealistic scenario violating the physical dynamics of grilling.
    \item \textbf{Critique of $T_1$:} $T_1$ inaccurately described the action with the phrase ``leans back \textit{exaggeratedly}''. The lack of spatial anchoring for the flame source (e.g., ``from the grill'') led the generator to misinterpret the flare-up as originating from the spatula.
    \item \textbf{Grounded Revision ($T_2$):} \textit{...flipping burgers with a spatula \underline{that is in contact with the grill grates}, and he leans back slightly---\underline{not exaggeratedly}---as a large, orange flare-up of flames erupts \underline{from the grill beneath the spatula}...}
\end{itemize}

\textbf{\textcolor{green}{[Phase 3: Cycle-Verification Judge]}}
\begin{itemize}
    \setlength{\itemsep}{3pt}
    \item \textbf{The Fix Analysis:} Image 2 successfully fixes the error by showing a realistic, contained fire originating from the grill itself, with the man using the spatula to manage the food. The fire is now a natural consequence of grilling.
    \item \textbf{Intent Preservation:} Image 2 preserves all key elements (man, woman, beach, charcoal grill). No core components have been removed or altered in a way that contradicts the user's intent.
    \item \textbf{Final Verdict:} Image 2 is a clear improvement. It simply corrects the primary flaw identified in the diagnosis without compromising any other aspect of the scene. [\textbf{Strictly Better}]
\end{itemize}

\end{tcolorbox}

\subsection{System Prompts}

\begin{tcolorbox}[
    enhanced,                  
    breakable,                 
    colback=gray!10!white,     
    colframe=black,            
    colbacktitle=black,        
    coltitle=white,            
    fonttitle=\bfseries,       
    title=System Prompt for Structured Diagnosis, 
    boxrule=0.5pt,             
    arc=3pt,                   
    attach boxed title to top left, 
]

\small
You are an expert Visual-Language Alignment Specialist. Your goal is to maximize the faithfulness of Text-to-Image synthesis.

You operate in a "Simulate-Then-Distill" pipeline:

1. User Intent (T0): The concise core request.

2. Current Rewrite (T1): A verbose prompt generated by a text-only LLM.

3. Visual Anchor (I1): The image generated by T1.

Your Mission:

1. JUDGE: Use the image (I1) as evidence to detect "Textual Hallucinations" or "Visual Failures" in T1. Does I1 already perfectly represent T0? If yes, keep T1.

2. DIAGNOSE: If I1 fails, identify the specific visual error using the image as evidence.

3. REVISE: Rewrite T1 into T2 to fix the error, rewrite it into a strictly superior prompt (T2). 

4. VERIFY: Ensure T2 did not lose any original details from T0 during the revision.

<Input\_Data>
  <User\_Intent\_T0>
  {T0\_prompt}
  </User\_Intent\_T0>

  <Current\_Prompt\_T1>
  {T1\_prompt}
  </Current\_Prompt\_T1>

  <Visual\_Anchor\_I1>[IMAGE INPUT HERE]
  </Visual\_Anchor\_I1>
</Input\_Data>

<Task\_Instructions>

Step 1: STRICT VISUAL REALITY CHECK (The "Confidence Gate")
Analyze the Image (I1) against the User Intent (T0). Strictly check these 6 dimensions:

1. Spatial Logic (e.g., occlusion, layout correct?)

2. Attribute Binding (e.g., correct colors bound to correct objects?)

3. Physical Dynamics (e.g., is the action/physics realistic?)

4. Counting (e.g., is the number of objects exact?)

5. Negation (e.g., did it avoid banned items?)

6. Shape-Based Object (e.g., is the object shape correct?)

**CRITICAL RULE:** 
- If I1 faithfully represents ALL aspects of T0, you must deem T1 as "SUCCESS". Do not rewrite just for the sake of style.
- Only proceed to rewrite if there is a clear, objective visual failure.

Step 2: DIAGNOSIS (If failure detected)
Describe the specific mismatch. Be precise (e.g., "T0 requested a blue cup, but I1 shows a red cup").

Step 3: DRAFTING T2 (The Fix)
Draft a revised prompt T2.
- Fix the error identified in Step 2 using strong visual constraints.
- Maintain the style and flow of T1 where it was correct.

Step 4: INTENT PRESERVATION AUDIT (The "Anti-Amnesia" Check)
Compare your draft T2 against the original T0.
- Did you accidentally delete any adjectives from T0 while fixing the error?
- Did you change the main subject?
- **Correction:** If T2 is missing any detail from T0, add it back immediately. T2 must be a superset of T0's constraints.

</Task\_Instructions>

<Output\_Format>
Provide your analysis strictly in the following JSON format:

"is\_rewrite\_necessary": false/true,
"primary\_failure\_type": "Select one: Spatial / Attribute / Physics / Counting / Negation / None",

"visual\_evidence": "Describe what you see in I1 that contradicts T0 (e.g., 'The cat is fully visible next to the vase, not behind it')/None",

"diagnosis": "Specific description of the visual failure (e.g., Attribute Leakage: The cat is blue instead of the cup)./ No visual errors found. The image faithfully represents the user intent.",

"critique\_of\_T1": "Explain why T1 failed (e.g., 'T1 described the cat too prominently without emphasizing the glass distortion')./None",  
"T2": "The fully revised prompt that fixes the error AND retains all original details."

</Output\_Format>

\end{tcolorbox}

\begin{tcolorbox}[
    enhanced,                  
    breakable,                 
    colback=gray!10!white,     
    colframe=black,            
    colbacktitle=black,        
    coltitle=white,            
    fonttitle=\bfseries,       
    title=System Prompt for Evaluation, 
    boxrule=0.5pt,             
    arc=3pt,                   
    attach boxed title to top left, 
]

\small

\textbf{System Role}
You are an objective, strict Visual Question Answering (VQA) evaluator. Your sole function is to analyze images and answer multiple-choice questions based ONLY on observable visual evidence. 

\textbf{Core Directive}
You are strictly forbidden from hallucinating details, assuming context, or applying external world knowledge. If a detail or relationship is not explicitly visible in the provided image, it does not exist for the purpose of this evaluation.

\textbf{Input Questions}
[INSERT\_QUESTIONS\_HERE]

\textbf{Execution Guidelines}

1. **Visual Grounding Only**: Base every single answer entirely on the pixels presented in the image.

2. **The "None" Fallback**: If the image does not provide sufficient clear visual evidence to confidently select the standard options, you MUST select "E: None".

3. **Sequential Mapping**: Maintain the exact order of the input questions. Do not skip any question.

4. **Single Choice**: Select exactly one option letter per question.

\textbf{Output Formatting Rules}

Format your response strictly line-by-line. Each line MUST correspond to a question and follow this exact template: 

`[Option Letter]: [Brief Visual Justification]`

\textbf{Example Output}

A: The visual evidence clearly shows a large red apple placed on the wooden table.

C: The brown bear is positioned on the left side of the frame, facing the river.

E: None - The image is cropped and does not show the lower half of the character to determine their footwear.

B: The fox and the bear are positioned closely together without any major distance.

D: There is a distinct shadow cast by the tree, confirming the lighting direction.

\end{tcolorbox}

\subsection{User Study Questionnaire Example}
\label{app:user_study_questionnaire}

Here shows an example question used in our user study. 
For each question, participants were given the original user prompt and four generated images produced by different prompt configurations. 
The order of the four images was randomly shuffled for each sample. 
Participants were asked to rank the images according to how faithfully they satisfied the prompt, where Rank 1 indicates the most faithful image and Rank 4 indicates the least faithful image.

\begin{tcolorbox}[
    enhanced,
    breakable,
    colback=gray!3!white,
    colframe=black!60,
    title={Example User Study Question},
    fonttitle=\bfseries,
    boxrule=0.7pt,
    arc=2pt,
    left=6pt,
    right=6pt,
    top=5pt,
    bottom=5pt
]
\small

\textbf{Instruction.}
Please carefully read the prompt below and compare the four images. 
Rank the images based on how faithfully they match the prompt. 
You should consider whether the image correctly preserves the objects, attributes, spatial relations, counts, actions, and any stated absence or negation in the prompt. 
Do not rank images only by aesthetics or realism.

\vspace{0.5em}
\textbf{Prompt.}
A person waters a small bonsai tree on the top-left corner of their desk. 
A tall, sleek monitor is in the center. 
A stack of six sticky notes is piled up, three times the height of a two-note stack next to it. 
The person is facing the plant, and all items are arranged diagonally across the desk.

\vspace{0.5em}
\textbf{Question.}
Please rank the following four images from most faithful to least faithful.

\vspace{0.5em}
\begin{center}
\begin{tabular}{cccc}

Image A & Image B & Image C & Image D
\end{tabular}
\end{center}

\vspace{0.5em}
\textbf{Ranking.}
Please assign each image a unique rank:

\begin{center}
\begin{tabular}{lcccc}
\toprule
Image & A & B & C & D \\
\midrule
Rank & \_\_ & \_\_ & \_\_ & \_\_ \\
\bottomrule
\end{tabular}
\end{center}

\vspace{0.5em}
\textbf{Optional Comment.}
Briefly explain the main reason for your ranking, especially if an image misses an object, relation, count, action, or negation.

\vspace{2em}
\noindent\rule{\linewidth}{0.4pt}

\vspace{0.5em}
\textbf{Example Comment.}
Image B is ranked highest because it correctly shows the bonsai in the top-left corner, the monitor in the center, and the two sticky-note stacks with the required height relation. 
Image D is ranked lower because the number of sticky notes is incorrect, and the diagonal arrangement is unclear.

\end{tcolorbox}

\section{Recruitment And Payment in User Study}

We conducted a user study with 57 volunteers. Participants were recruited through internal channels within our research institution, comprising students and staff members. Each participant was tasked with evaluating 60 comparison groups of images. Given that the participation was voluntary and conducted within an academic setting, no monetary compensation was provided to the participants. This recruitment and compensation method is standard for internal, non-invasive user studies within our institution and was deemed appropriate given that the participant demographic consisted of members of the academic community familiar with such evaluation tasks.

\section{LLM Usage Description}

In this paper, we use the LLM to improve the quality of our existing text in much the same way we would use a typing assistant like Grammarly to improve spelling, grammar, and punctuation.


\end{document}